\documentclass{article}

\usepackage{times}
\usepackage{graphicx} % more modern
\usepackage{subfig}

\graphicspath{{./figs/}}

\usepackage{natbib}

\usepackage{algorithm}
\usepackage{algorithmic}

\usepackage{hyperref}

\usepackage[accepted]{icml2017}

\icmltitlerunning{DeepBach: a Steerable Model for 
  Bach Chorales Generation}

\begin{document} 

\twocolumn[
\icmltitle{DeepBach: a Steerable Model for 
  Bach Chorales Generation
}


% you can specify symbols, otherwise they are numbered in order
\icmlsetsymbol{equal}{*}

\begin{icmlauthorlist}
\icmlauthor{Ga\"etan Hadjeres}{upmc,sony}
\icmlauthor{Fran\c{c}ois Pachet}{upmc,sony}
\icmlauthor{Frank Nielsen}{sonyjap}
\end{icmlauthorlist}

\icmlaffiliation{upmc}{LIP6, Universit\'e Pierre et Marie Curie}
\icmlaffiliation{sony}{Sony CSL, Paris}
\icmlaffiliation{sonyjap}{Sony CSL, Japan}
%\icmlaffiliation{lix}{Computer Science Department LIX, École Polytechnique}

\icmlcorrespondingauthor{Ga\"etan Hadjeres}{gaetan.hadjeres@etu.upmc.fr}
\icmlcorrespondingauthor{Fran\c{c}ois Pachet}{pachetcsl@gmail.com}
\icmlcorrespondingauthor{Frank Nielsen}{Frank.Nielsen@acm.org}

\icmlkeywords{Automatic Music Generation, Bach Chorales}

\vskip 0.3in
]



\printAffiliationsAndNotice{}  % leave blank if no need to mention equal contribution
%\footnotetext{hi}

%break url after -
\expandafter\def\expandafter\UrlBreaks\expandafter{\UrlBreaks%  save the current one
  \do\-}

\begin{abstract}
  This paper introduces DeepBach, a graphical model aimed at modeling polyphonic music and specifically hymn-like pieces. 
We claim that, after being trained on the chorale harmonizations by Johann Sebastian Bach, our model is capable of
generating highly convincing chorales in the style of Bach.
DeepBach's strength comes from the use of pseudo-Gibbs sampling coupled with an adapted representation of musical data.
This is in contrast with many automatic music composition approaches  which tend to compose music sequentially. Our model is also steerable in the sense that a user can constrain the generation by imposing positional constraints such as notes, rhythms or cadences in the generated score.
We also provide a plugin on top of the MuseScore music editor making the interaction with DeepBach easy to use.
\end{abstract}

\section{Introduction}
\label{sec:introduction}

\begin{figure*}

  \centering
    \subfloat[Original text and melody by Georg Neumark (1641), ]{
      \includegraphics[width=0.40\textwidth, height=2.4cm]{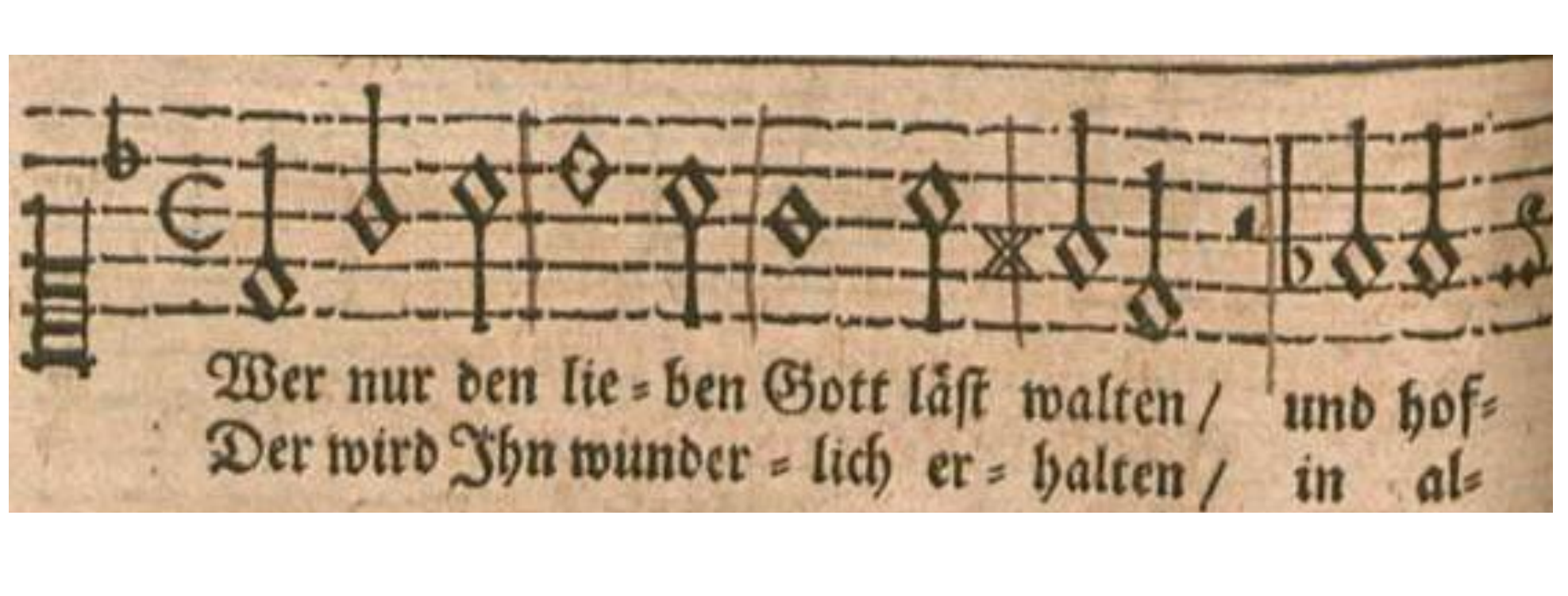}
      \label{bwv434original}

      }
\hfill
  \subfloat[Four-voice harmonization by Bach: voices are determined by the staff they are written on and the directions of the stems.]{
    \includegraphics[width=0.55\textwidth, height=2.3cm]{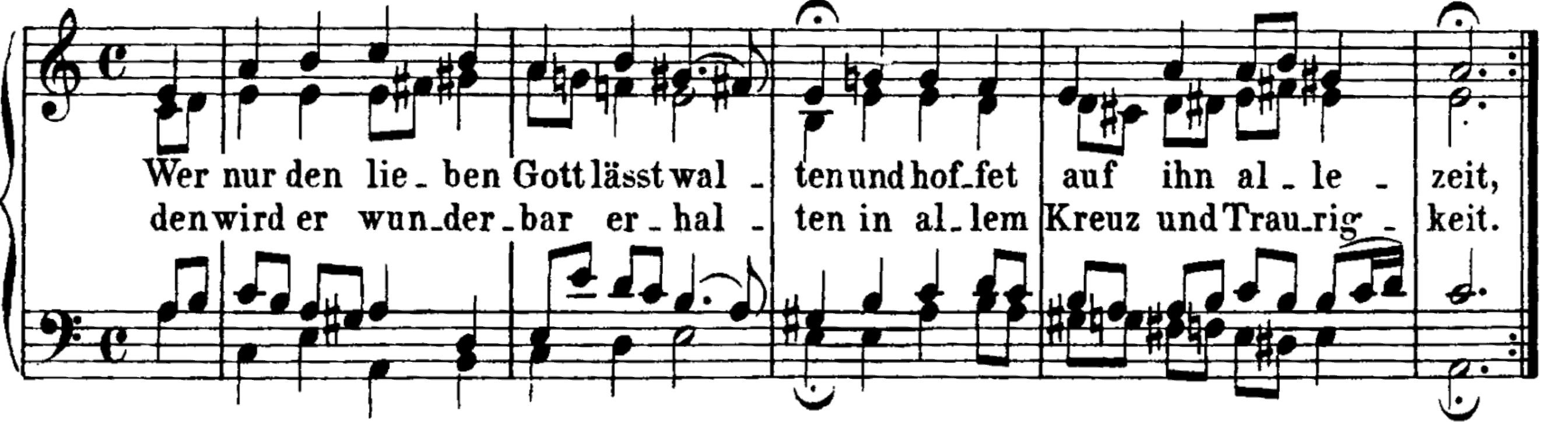}
     \label{fig:bwv434bach}
    
}

  \caption[BWV434 caption]{Two versions of ``Wer nur den lieben Gott l\"a{\ss}t walten''.  The original melody (a) and its
    reharmonization (b) by Johann Sebastian Bach (BWV 434) \footnote{bwv434}.}
  \label{reharmoExample}
\end{figure*}

The composition of polyphonic chorale music in the style of J.S. Bach has represented a major challenge in automatic music composition over the last decades.
%The art of Bach chorales composition involves combining four-part harmony with characteristic rhythmic patterns and
%typical melodic movements to produce musical phrases which begin, evolve and end (cadences) in a harmonious way.
% To our knowledge, no model so far was able to solve all these problems simultaneously using an agnostic machine-learning
% approach.
  The corpus of the chorale harmonizations by Johann Sebastian Bach  is remarkable by its homogeneity and its size (389 chorales in \cite{bach1985389}). All these short pieces (approximately one minute long) are written for a four-part chorus (soprano, alto, tenor and bass) using similar compositional principles: the composer takes a well-known (at that time) melody from a Lutheran hymn and harmonizes it i.e. the three lower parts (alto, tenor and bass) accompanying the soprano (the highest part) are composed, see Fig.\ref{reharmoExample} for an example.

Moreover, since the aim of reharmonizing a melody is to give more power or new insights to its text, the lyrics have to
be understood clearly. We say that voices are in \emph{homophony}, i.e. they articulate syllables simultaneously. This
implies characteristic  rhythms, variety of harmonic ideas as well as characteristic  melodic movements which make the
style of these chorale compositions easily distinguishable, even for non experts.

%footnote of fig:bwv434bach
\footnotetext{\url{https://www.youtube.com/watch?v=73WF0M99vlg}}
 The difficulty, from a compositional point of view comes from the intricate interplay between harmony (notes sounding
 at the same time) and voice movements (how a single voice evolves through time). Furthermore, each voice has its own
 ``style'' and its own coherence. Finding a chorale-like reharmonization which combines Bach-like harmonic progressions
 with musically interesting melodic movements is a problem which often takes years of practice for musicians.
 %Lots of  theoretical books have been produced by experts to assess what can be the correct progressions in the style
 %of Bach and which are not \cite{koechlin1928traite,benward2014music}.

 From the point of view of automatic music generation, the first solution to this
 apparently highly combinatorial problem  was proposed by \cite{ebcioglu1988} in 1988. This problem is seen as a
 constraint satisfaction problem, where the system must fulfill numerous hand-crafted constraints characterizing the
 style of Bach. It is a rule-based expert system which contains no less than 300 rules and tries to reharmonize a given
 melody with a generate-and-test method and intelligent backtracking. Among the short examples presented at the end of
 the paper, some are flawless. The drawbacks of this method are, as stated by the author, the considerable effort to
 generate the rule base and the fact that the harmonizations produced ``do not sound like Bach, except for
 occasional Bachian patterns and cadence formulas.'' In our opinion, the requirement of an expert knowledge  implies a
 lot of subjective choices.
 %Furthermore, we have no idea about the variety and originality of the proposed solutions.
 
 % As suggested in \cite{ebcioglu1988},
 A neural-network-based solution was later developed by \cite{hild1992harmonet}. This
 method relies on several neural networks, each one trained for solving a specific task:  a harmonic skeleton is first
 computed then refined and ornamented.  A similar approach is   adopted in \cite{allan2005harmonising}, but uses Hidden
 Markov Models (HMMs) instead of neural networks. Chords are represented as lists of intervals and form the states of
 the Markov models. These approaches produce interesting results even if they both use
 expert knowledge and bias the generation by imposing their compositional process. In
 \cite{whorley2013multiple,whorley2016}, authors elaborate on those methods by introducing multiple viewpoints and
 variations on the sampling method (generated sequences which violate ``rules of harmony'' are put aside for
 instance). However, this approach does not produce a convincing chorale-like texture, rhythmically as well as
 harmonically and the resort to hand-crafted criteria to assess the quality of the generated sequences might rule out
 many musically-interesting solutions.
 
Recently, agnostic approaches (requiring no knowledge
 about harmony, Bach's style or  music) using neural networks have been investigated with
promising results. In \cite{boulanger2012modeling}, chords are modeled with Restricted Boltzmann Machines (RBMs).  Their
temporal dependencies are learned using Recurrent Neural Networks (RNNs).
Variations of these architectures based on Long Short-Term Memory (LSTM) units (\cite{hochreiter1997long,mikolov2014learning}) or GRUs (Gated Recurrent Units) have been developed by \cite{Lyu2015} and
 \cite{chung2014empirical} respectively. However, these models which work on piano roll representations of the
music  are too general to capture the specificity of Bach chorales. Also, a major drawback is their lack of flexibility. Generation is performed from left to right. A
user cannot interact  with the system: it is impossible to do reharmonization for instance which is the essentially how
the corpus of Bach chorales was composed. Moreover, their invention capacity and non-plagiarism abilities are not
demonstrated.

A method that addresses the rigidity of sequential generation in music was first proposed in \cite{sakellariou:15a, 2016arXiv161003414S} for monophonic music and later generalized to polyphony in \cite{hadjeres2016style}. These approaches advocate for the use of Gibbs sampling as a generation process in automatic music composition.
%Another similar approach based on blocked-Gibbs sampling was described in \cite{huang}. Their method is based on a convolutional neural network trained at reconstructing a masked portion of a piano-roll.

The most recent advances in chorale harmonization is arguably the BachBot model \cite{liang2016bachbot}, a
LSTM-based approach specifically designed to deal with Bach chorales. This approach relies on little musical knowledge
(all chorales are transposed in a common key) and is able to produce high-quality chorale harmonizations. However,
compared to our approach, this model is less general (produced chorales are all in the C key for instance) and less
flexible (only the soprano can be fixed).
Similarly to our work, the authors evaluate their model
with an online Turing test to assess the efficiency of their model.
They also take into account
the fermata symbols (Fig.~\ref{fig:fermata}) which are indicators of the structure of the chorales.

In this paper we introduce DeepBach, a dependency network \cite{heckerman2000dependency} capable of producing musically convincing four-part chorales in
the style of Bach by using a Gibbs-like sampling procedure.  
Contrary to models based on RNNs, we do not sample from left to right which  allows us to enforce positional, unary user-defined constraints such as rhythm, notes, parts, chords and
cadences. DeepBach is able to generate coherent musical phrases and provides, for instance, varied reharmonizations of melodies without plagiarism. Its core features are its speed, the possible interaction
with users and the richness of harmonic ideas it proposes. Its efficiency opens up new ways of composing
Bach-like chorales for non experts in an interactive manner similarly to what is proposed in \cite{Papadopoulos2016} for leadsheets.

In Sect.~\ref{sec:deepbach} we present the DeepBach model for four-part chorale generation.
We discuss  in Sect.~\ref{sec:experiments} the results of an experimental study we conducted to assess the quality of
our model. Finally, we provide generated examples in Sect.~\ref{sec:examples} and elaborate on the possibilities offered by our interactive music composition editor in Sect.~\ref{sec:interactive}. All examples can be heard on the
accompanying web page\footnote{\url{https://sites.google.com/site/deepbachexamples/}} and the
code of our implementation is available on GitHub\footnote{\url{https://github.com/Ghadjeres/DeepBach}}. Even if our presentation focuses on Bach chorales, this model has been successfully applied to other styles and composers including Monteverdi five-voice madrigals to Palestrina masses.

\section{DeepBach}
\label{sec:deepbach}
In this paper we introduce a  generative model which takes into account the distinction between
voices.
Sect.~\ref{sec:data-representation} presents the data representation we used. This representation is both fitted for our sampling procedure and more accurate than many data representation  commonly  used in automatic music composition.
Sect.~\ref{sec:model-architecture} presents the model's architecture and Sect.~\ref{sec:generation} our generation method. Finally, Sect.~\ref{sec:impl-deta}  provides implementation details and indicates how we preprocessed the corpus of Bach chorale harmonizations.

\subsection{Data Representation}
\label{sec:data-representation}

\subsubsection{Notes and Voices}
We use MIDI pitches to encode notes and choose to model voices separately. We consider that only one note can be sung at a given time and discard chorales with voice divisions.

Since Bach chorales only contain simple time signatures, we discretize time with sixteenth notes, which means that each beat is subdivided into four equal parts. Since there is no smaller subdivision in Bach chorales, there is no loss of information in this process. 

In this setting, a voice $\mathcal{V}_i = \{ \mathcal{V}_i^t \}_t$ is a list of notes indexed by $t \in [T]\footnote{We adopt the standard notation $[N]$ to denote the set of integers $\{1, \dots, N\}$ for any integer $N$.}$, where $T$ is the duration piece (in sixteenth notes).

\subsubsection{Rhythm}
\label{sec:rhythm}

We choose to model rhythm by simply adding a \emph{hold symbol} ``\_\_'' coding whether or not the preceding note is held to the list of existing notes. This representation is thus unambiguous, compact and  well-suited  to our sampling method (see Sect.~\ref{sec:import-data-repr}).

\subsubsection{Metadata}
\label{sec:metadata}
\begin{figure}[h]
  \centering
    \includegraphics[scale=0.3]{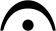}
\caption{Fermata symbol.}
  \label{fig:fermata}
\end{figure}

The music sheet (Fig.~\ref{fig:bwv434bach}) conveys  more information than only the notes played. We can cite:
\begin{itemize}
\item the lyrics,
\item the key signature,
\item the time signature,
\item the beat index,
\item an implicit metronome (on which subdivision of the beat the note is played),
\item the fermata symbols (see Fig.~\ref{fig:fermata}),
\item current key,
\item current key  signature,
\item current mode (major/minor/dorian).
\end{itemize}

In the following, we will only take into account the fermata symbols, the subdivision indexes and the current key signature. To this end, we introduce:
\begin{itemize}
\item The \emph{fermata} list 
$\mathcal{F}$ that indicates if there is a fermata symbol, see Fig.~\ref{fig:fermata}, over the current note, it is a Boolean
value. If a fermata is placed over a note on the music sheet, we consider that it is active for all time indexes within
the duration of the note.

\item The \emph{subdivision} list
  $\mathcal{S}$  that contains the subdivision indexes of the beat. It is an integer between 1 and 4: there is no distinction between beats in a bar so that our model is able to deal with chorales with three and four beats per measure.
  
\end{itemize}

\subsubsection{Chorale}
\label{sec:chorale}
We represent a chorale as a couple
\begin{equation}
  \label{eq:1}
(\mathcal{V}, \mathcal{M})  
\end{equation}
composed  of voices and metadata. For Bach chorales, $\mathcal{V}$ is a list of 4 voices $\mathcal{V}_i$ for $i \in [4]$  (soprano, alto, tenor and bass) and $\mathcal{M}$ a collection of metadata lists ($\mathcal{F}$ and $\mathcal{S}$).

\begin{figure}
%\quad \quad 
  \subfloat[]{

    \includegraphics[width=0.3\linewidth, height=2.1cm, angle=0.5, trim=0 0 0 0, clip=true]{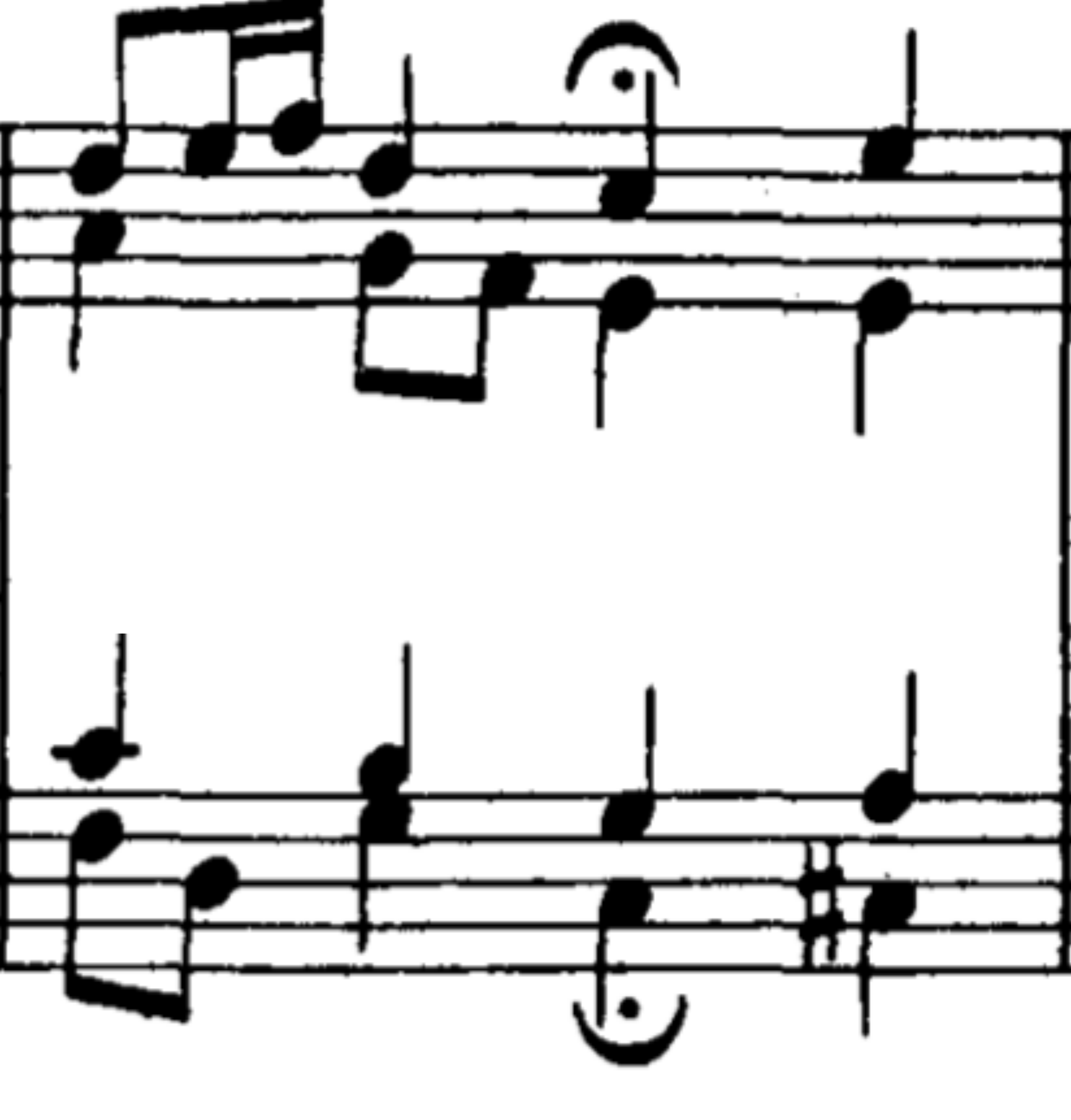}
    \label{fig:extract}
  }
  \subfloat[]{
        \includegraphics[width= 0.65 \linewidth, trim=0 100 0 100]{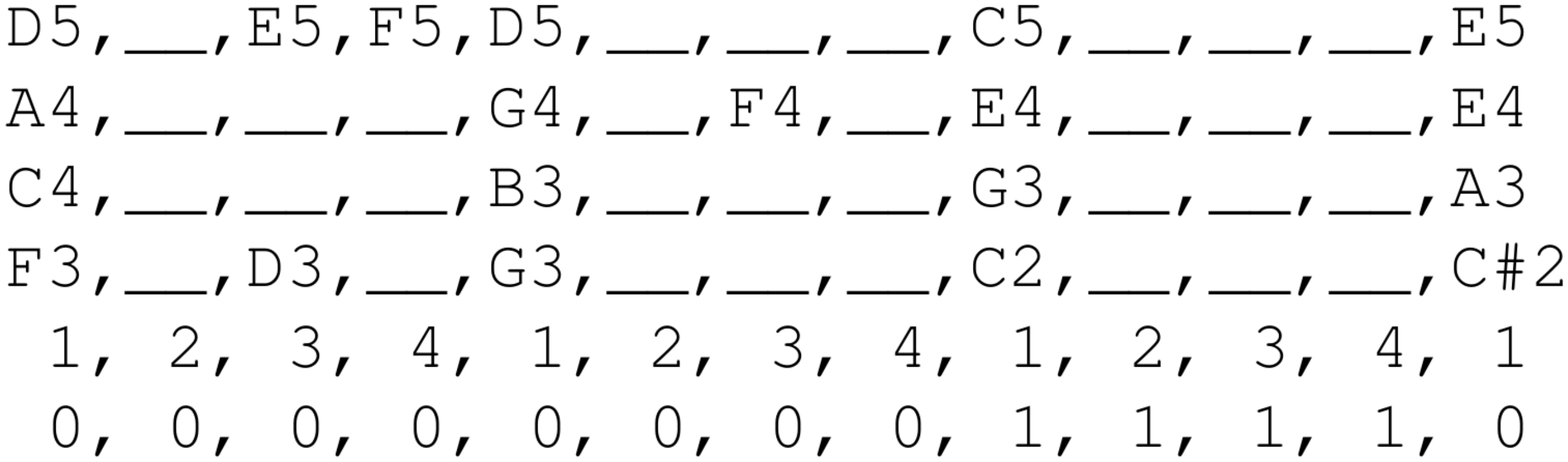}
        \label{fig:input}
    }
%    \caption{Chorale extract }
  \caption{Extract from a Bach chorale and its representation as four voice lists and two metadata lists ($\mathcal{S}$ and $\mathcal{F}$). The hold symbol is displayed as ``\_\_'' and considered as a note.}
  \end{figure}
  % \quad

% \begin{verbatim}
% [ D5,__,E5,F5,D5,__,__,__,C5,__,__,__,E5 ,__,__,__]
% [ A4,__,__,__,G4,__,F4,__,E4,__,__,__,E4 ,__,__,__]
% [ C4,__,__,__,B3,__,__,__,G3,__,__,__,A3 ,__,__,__]
% [ F3,__,D3,__,G3,__,__,__,C2,__,__,__,C#2,__,__,__]
% [  1, 2, 3, 4, 1, 2, 3, 4, 1, 2, 3, 4, 1 , 2, 3, 4]
% [  0, 0, 0, 0, 0, 0, 0, 0, 1, 1, 1, 1, 0 , 0, 0, 0]
% \end{verbatim}

  Our choices are very general and do not involve expert knowledge about harmony or scales but are only mere
  observations of the corpus. The list $\mathcal{S}$ acts as a metronome. The list $\mathcal{F}$ is added since fermatas
  in Bach chorales indicate the end of each musical phrase. The use of fermata to this end is a specificity of Bach
  chorales that we want to take advantage of.
  %Part~\ref{sec:examples} shows that this representation makes our model   able to create convincing musical phrases in triple and quadruple simple time signatures.

\subsection{Model Architecture}
\label{sec:model-architecture}
We choose to consider the metadata sequences in $\mathcal{M}$  as given.
For clarity, we suppose in this section that our dataset is composed of only one chorale written as in Eq.~\ref{eq:1} of size $T$. We define a \emph{dependency network} on the finite set of variables $\mathcal{V} = \{V_i^t\}$ by specifying a set of conditional probability distributions (parametrized by parameter $\theta_{i, t}$)
\begin{equation}
  \label{conditionals}
  \left\{ p_{i,t}(V_i^t | V_{\setminus i,t},  \mathcal{M}, \theta_{i, t}) \right\}_{ i \in [4], t \in [T]} ,
\end{equation}
where  $\mathcal{V}_i^t$ indicates the note of voice $i$ at time index $t$ and  $\mathcal{V}_{\setminus i,t}$ all variables in $\mathcal{V}$ except from the variable $\mathcal{V}_i^t$.
As we want  our model to be time invariant so that we can apply it to sequences of any size, we share the parameters between all conditional probability distributions on variables lying in the same voice, i.e. \[\theta_i := \theta_{i,t}, \quad p_i := p_{i,t} \quad \forall t \in [T].\]

Finally, we fit each of these conditional probability distributions on the data by maximizing the log-likelihood. Due to weight sharing, this amounts to solving
% \begin{equation}
%   \label{eq:pseudolikelihood}
%   -\sum_i \left(\sum_t \log p(\mathcal{V}_i^t| \mathcal{V}_{\setminus i,t}, \mathcal{M}, \theta_i) \right),
% \end{equation}
%
%We consider only ``local'' interactions between notes, which means that 
%
%We need to find a parameter $\theta$ which minimizes this loss. In order to have a computationally tractable training criterion, we introduce the \emph{pseudolikelihood} of our data \cite{besag1975statistical,arnold1991pseudolikelihood}. This approach proved successful in many real-life problems \cite{ekeberg2013improved} or music generation \cite{2016arXiv161003414S, hadjeres2016style}
%and  consists in an approximation of the negative log-likelihood function by the sum over all variables: 
%
%where This suggests to introduce four probabilistic models $p_i$ depending on parameter $\theta_i$, one for each voice, and to minimize their negative log-likelihood  independently using the pseudolikelihood criterion.
 four classification problems of the form:
\begin{equation}
  \label{eq:pseudolikelihoodVoice}
\max_{\theta_i} \sum_t \log p_i(\mathcal{V}_i^t| \mathcal{V}_{\setminus i,t}, \mathcal{M}, \theta_i), \quad  \textrm{ for } i \in [4],        
\end{equation}
where the aim is to predict a note knowing the value of its neighboring notes, the subdivision of the beat it is on and the presence of fermatas.
The advantage with this formulation is that each classifier has to make predictions within a small range of notes whose ranges correspond to the notes within the usual voice ranges (see \ref{sec:impl-deta}). 

\begin{figure}                                                                                                          \centering
        \includegraphics[width=0.7\linewidth]{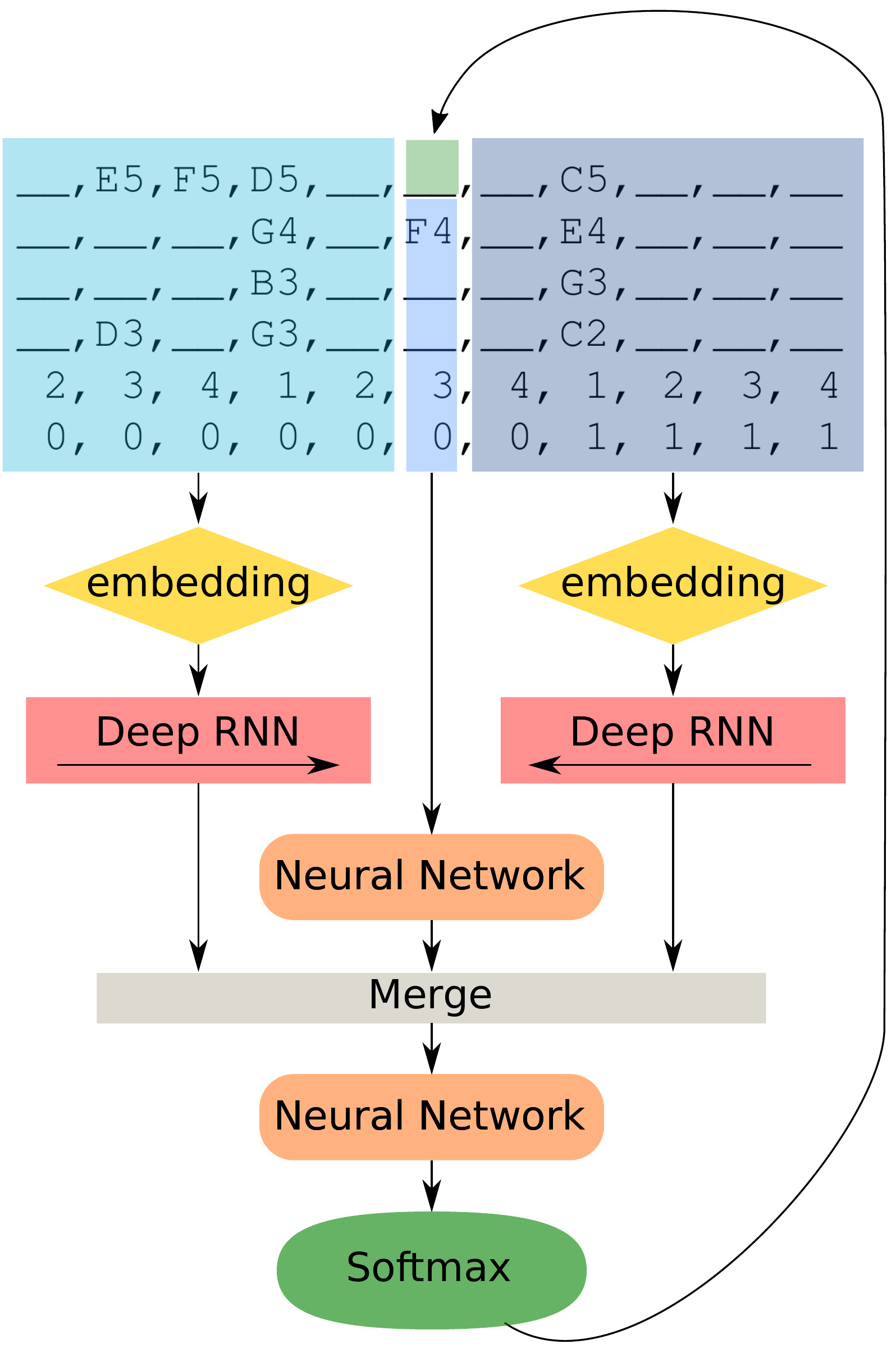}
\caption{Graphical representations of DeepBach's neural network architecture for the soprano prediction $p_1$.}
  \label{fig:deepbach-architecture}
\end{figure}

For accurate predictions and in order to take into account the sequential aspect of the data, each classifier is modeled  using four neural networks: two Deep Recurrent Neural Networks \cite{2013arXiv1312.6026P}, one summing up \emph{past} information and another summing up information coming from the \emph{future} together with a non-recurrent neural network for notes occurring at the same time. Only the last output from the uppermost  RNN layer is kept. These three outputs are then merged and passed as the input of a fourth neural network whose output is  $p_i(\mathcal{V}_i^t| \mathcal{V}_{\setminus i,t}, \mathcal{M}, \theta)$. Figure~\ref{fig:deepbach-architecture} shows a graphical representation for one of these models. Details are provided in Sect.~\ref{sec:impl-deta}. These choices of architecture somehow match real compositional practice on Bach chorales. Indeed, when reharmonizing a given melody, it is often simpler to start from the cadence and write music ``backwards.''

\subsection{Generation}
\label{sec:generation}
\subsubsection{Algorithm}
Generation in dependency networks is performed using the pseudo-Gibbs sampling procedure. This Markov Chain Monte Carlo (MCMC) algorithm is described in Alg.\ref{alg:1}. It is similar to the classical Gibbs sampling procedure \cite{geman1984stochastic} on the difference that the conditional distributions are potentially incompatible \cite{doi:10.1080/00949655.2014.968159}. This means that the conditional distributions of Eq.~(\ref{conditionals}) do not necessarily comes from a joint distribution $p(\mathcal{V})$ and that the theoretical guarantees that the MCMC converges to this stationary joint distribution vanish. We experimentally verified that it was indeed the case by checking that the Markov Chain of Alg.\ref{alg:1} violates Kolmogorov's criterion \cite{kelly2011reversibility}: it is thus not reversible and cannot converge to a joint distribution whose conditional distributions match the ones used for sampling.

However, this Markov chain converges to another stationary distribution and applications on real data demonstrated that this method yielded accurate joint probabilities, especially when the inconsistent probability distributions are learned from data \cite{heckerman2000dependency}. Furthermore, nonreversible MCMC algorithms can in particular cases be better at sampling that reversible Markov Chains \cite{2014arXiv1412.8762V}.

\begin{algorithm}
  
  \begin{algorithmic}[1]
    \STATE {\bfseries Input:} {Chorale length $L$, metadata $\mathcal{M}$ containing lists of length $L$, probability distributions $(p_1, p_2, p_3, p_4)$, maximum number of iterations $M$}

    \STATE Create four lists $\mathcal{V} = (\mathcal{V}_1, \mathcal{V}_2, \mathcal{V}_3, \mathcal{V}_4)$ of length $L$
    \STATE
    \COMMENT{The lists are  initialized with random notes drawn from the ranges of the corresponding voices (sampled uniformly or from the marginal distributions of the notes)}
          
    \FOR{$m$ from $1$ to $M$ }
    \STATE Choose voice $i$ uniformly between 1 and 4
    \STATE Choose time $t$ uniformly between 1 and  $L$
    \STATE Re-sample $\mathcal{V}_i^t$ from $p_i(\mathcal{V}_i^t| \mathcal{V}_{\setminus i,t}, \mathcal{M}, \theta_i)$
    \ENDFOR
    \STATE {\bfseries Output:} $\mathcal{V} = (\mathcal{V}_1, \mathcal{V}_2, \mathcal{V}_3, \mathcal{V}_4)$
\end{algorithmic}
\caption{Pseudo-Gibbs sampling}
\label{alg:1}
\end{algorithm}

\subsubsection{Flexibility of the  sampling procedure}
The advantage of this method is that we can enforce user-defined constraints by tweaking Alg.~\ref{alg:1}:
\begin{itemize}
\item instead of choosing voice $i$ from $1$ to $4$ we can choose to fix the soprano and only resample voices from $2$, $3$ and $4$ in step (3) in order to provide reharmonizations of the fixed melody
\item we can choose the fermata list $\mathcal{F}$ in order to impose end of musical phrases at some places
\item more generally, we can impose any metadata
\item for any $t$ and any $i$, we can fix specific subsets $\mathcal{R}_i^t$  of notes  within the range of voice $i$. We then restrict ourselves to some specific chorales by re-sampling  $\mathcal{V}_i^t$ from \[p_i(\mathcal{V}_i^t| \mathcal{V}_{\setminus i,t}, \mathcal{M}, \theta_i,\mathcal{V}_i^t \in \mathcal{R}_i^t)\] at step (5). This allows us for instance to fix rhythm (since the hold symbol is considered as a note), impose some chords in a soft manner or restrict the vocal ranges.
\end{itemize}

\subsubsection{Performance}
Note that it is possible to make generation faster by making parallel Gibbs updates on GPU. Steps (3) to (5) from
Alg.~\ref{alg:1} can be run simultaneously to provide significant speedups.
%In Table~\ref{tab:1} we show how the batch size
%(fixed number of parallel updates) influences the number of updates per second.
Even if it is known that this approach is
biased \cite{de2016ensuring} (since we can update simultaneously variables which are not conditionally independent), we experimentally observed that for small batch sizes ($16$ or $32$), DeepBach still
generates samples of great musicality while running ten times faster than the sequential version. This
allows DeepBach to generate chorales in a few seconds.

% \begin{table}
%   \centering
% \begin{tabular}{c|c}
%   Batch size &   Number of updates per second \\
%   \hline
%   1 & 59 \\
%   2 & 112 \\
%   4 & 200 \\
%   8 & 334 \\
%   16 & 508 \\
%   32 & 636 \\
%   64 & 693 \\
%   128 & 824 \\
%   256 & 914 \\
%   512 & 974 \\
%   1024 & 1017 \\
% \end{tabular}
% \caption{Mean number of Gibbs updates per second during DeepBach's generation as a function of the batch size  using a Nvidia
%   GTX 980Ti GPU.}
% \label{tab:1}
% \end{table}

It is also possible to use the hard-disk-configurations generation algorithm (Alg.2.9 in  \cite{krauth2006statistical}) to appropriately choose all the time indexes at which we parallelly resample so that:
\begin{itemize}
\item every time index is at distance at least $\delta$ from the other time indexes
  
\item configurations of time indexes satisfying the relation above are equally sampled.
\end{itemize}
This trick allows to assert that we do not update simultaneously a variable and its local context.

\subsubsection{Importance of the data representation}
\label{sec:import-data-repr}
We emphasize on this section the importance of our particular choice of data representation with respect to our sampling procedure. The fact that we obtain great results using  pseudo-Gibbs sampling relies exclusively on our choice to integrate the hold symbol into the list of notes.

Indeed, Gibbs sampling fails to sample the true joint distribution $p(\mathcal{V}| \mathcal{M}, \theta)$ when variables are highly correlated, creating  isolated regions of high probability states in which the MCMC chain can be trapped.  However, many data representations used in music modeling such as
\begin{itemize}
\item  the piano-roll representation,
  
\item the couple (\emph{pitch}, \emph{articulation}) representation where \emph{articulation} is a Boolean value indicating whether or not the note is played or held,
\end{itemize}
tend to make the musical data suffer from this drawback.

As an example, in the piano-roll representation, a long note is represented as the repetition of the same value over many variables. In order to only change its pitch, one needs to change simultaneously a large number of variables (which is exponentially rare) while this is achievable with only one variable change with our representation.

\subsection{Implementation Details}
\label{sec:impl-deta}

We implemented DeepBach using Keras \cite{chollet2015keras} with the Tensorflow \cite{tensorflow2015-whitepaper}
backend. We used the database of chorale harmonizations by J.S. Bach included in the music21 toolkit \cite{cuthbert2010music21}.
After removing chorales with instrumental parts and chorales containing parts with two simultaneous notes (bass
parts sometimes divide for the last chord), we ended up with 352 pieces. Contrary to other approaches which transpose
all chorales to the same key (usually in C major or A minor), we choose to augment our dataset by adding all chorale
transpositions which fit within the vocal ranges defined by the initial corpus. This gives us a corpus of 2503 chorales
and split it between a training set (80\%) and a validation set (20\%).
 The vocal ranges contains less than 30
different pitches for each voice (21, 21, 21, 28) for the soprano, alto, tenor and bass parts respectively.

As shown in Fig.~\ref{fig:deepbach-architecture}, we model only \emph{local} interactions between a note $\mathcal{V}_i^t$ and its context ($\mathcal{V}_{\setminus i,t}$,  $\mathcal{M}$) i.e. only elements with time index $t$ between $t - \Delta t$ and $t + \Delta t$ are taken as inputs of our model for some scope $\Delta t$. This approximation appears to be accurate since  musical analysis reveals that  Bach chorales do not exhibit clear long-term dependencies. 

The reported results in Sect.~\ref{sec:experiments} and examples in  Sect.~\ref{sec:examples} were obtained with
$\Delta t=16$. We chose as the ``neural network brick'' in Fig.~\ref{fig:deepbach-architecture} a neural network with
one hidden layer of size 200 and ReLU \cite{nair2010rectified} nonlinearity and as the ``Deep RNN brick'' two stacked LSTMs \cite{hochreiter1997long,mikolov2014learning}, each one being of size 200 (see Fig.~2 (f) in \cite{li2015constructing}). The ``embedding brick'' applies the same neural network to each time slice $(\mathcal{V}_t, \mathcal{M}_t)$. There are 20\%  dropout on input and 50\% dropout after each layer.

We experimentally found
that sharing weights between the left and right embedding layers improved neither validation accuracy nor the musical quality of our generated chorales.

\section{Experimental Results}
\label{sec:experiments}
We evaluated the quality of our model with an online test conducted on human listeners.

\subsection{Setup}
\label{sec:test}
For the parameters used in our experiment, see Sect~\ref{sec:impl-deta}. We compared our model with two other models:
a Maximum Entropy model (MaxEnt) as in \cite{hadjeres2016style} %(Fig.~\ref{fig:maxent-architecture})
and a Multilayer
Perceptron (MLP) model. %(Fig.~\ref{fig:mlp-architecture}).

The Maximum Entropy model is a  neural network with no hidden layer. It is given by:
\begin{equation}
  \label{eq:maxent}
  p_i(\mathcal{V}_i^t| \mathcal{V}_{\setminus i,t}, \mathcal{M}, A_i, b_i) = \textrm{Softmax}(AX + b)
\end{equation}
where $X$ is a vector containing the elements in $\mathcal{V}_{\setminus i,t} \cup \mathcal{M}_t$,
$A_i$ a $(n_i,m_i)$ matrix and $b_i$ a vector of size $m_i$ with $m_i$ being the size of $X$, $n_i$ the number of
notes in the voice range $i$ and $\textrm{Softmax}$ the softmax function given by
\[\textrm{Softmax}(z)_j = \frac{e^{z_j}}{\sum_{k=1}^K e^{z_k}} \quad \textrm{ for } j \in [K] ,\]
for a vector $z = (z_1, \dots, z_K)$.

The Multilayer Perceptron model we chose takes as input elements in
$\mathcal{V}_{\setminus i,t} \cup \mathcal{M}$, is a neural network with one hidden layer of size 500 and uses a ReLU \cite{nair2010rectified}  nonlinearity.

All models are local and have the same scope $\Delta t$, see Sect.~\ref{sec:impl-deta}.

Subjects were asked to give information about their musical expertise. They could choose what category fits them best between:
\begin{enumerate}
\item I seldom listen to classical music
\item Music lover or musician
\item Student in music composition or professional musician.
\end{enumerate}

The musical extracts have been obtained by reharmonizing 50 chorales from the validation set by each of the three
models (MaxEnt, MLP, DeepBach). We rendered the MIDI files using the Leeds Town Hall Organ soundfont\footnote{\url{https://www.samplephonics.com/products/free/sampler-instruments/the-leeds-town-hall-organ}} and cut two extracts of
12 seconds from each chorale, which gives us 400 musical extracts for our test: 4 versions for each of the 100 melody
chunks. We chose our rendering so that the generated parts (alto, tenor and bass) can be distinctly heard and
differentiated from the soprano part (which is fixed and identical for all models): in our mix, dissonances are easily
heard, the velocity is the same for all notes as in a real organ performance and the sound does not decay, which is
important when evaluating the reharmonization of long notes.

% \subsubsection{Perception Test}
% \label{sec:perception-test}
% In a first part, subjects were presented ten series of two reharmonizations of the \emph{same} chorale melody and were
% asked ``which one sounds more like Bach to your ears''. In order to give a general ranking from these binary
% confrontations, we used the Bradley-Terry model \cite{bradley1952,Agresti2011} to infer potentials $\beta_j$ reflecting
% the probability that the version $j$ is better than another version. This is expressed as:
% \begin{equation}
%   \label{eq:bradley}
%   P(\textrm{version } i \textrm{ is better than version } j) = \frac{e^{\beta_i}}{e^{\beta_i} + e^{\beta_j}}.
% \end{equation}

% Results are plotted in Fig.~\ref{votesResults}.

% 1609 people took this test, 395 with musical expertise 1, 792 with musical expertise 2 and 422 with musical expertise 3.
% \begin{figure}[]

%     \centering
%     \includegraphics[width=\linewidth]{terryReshapedV4.pdf}

%     \caption{Results of the perception test. The figure shows, for each level of expertise (from 1 to 3), the potentials
%       of the Bradley-Terry model obtained from the pairwise comparisons. We centered the plots since only the distance
%       between points matters. Better seen in color.}
% \label{votesResults}
% \end{figure}

% Extracts generated from DeepBach are clearly recognized as being more Bach-like than the other models. The more musical expertise subjects have, the clearer is the distinction.

\begin{figure}[t]

    \centering
    \includegraphics[angle=0, width=\linewidth]{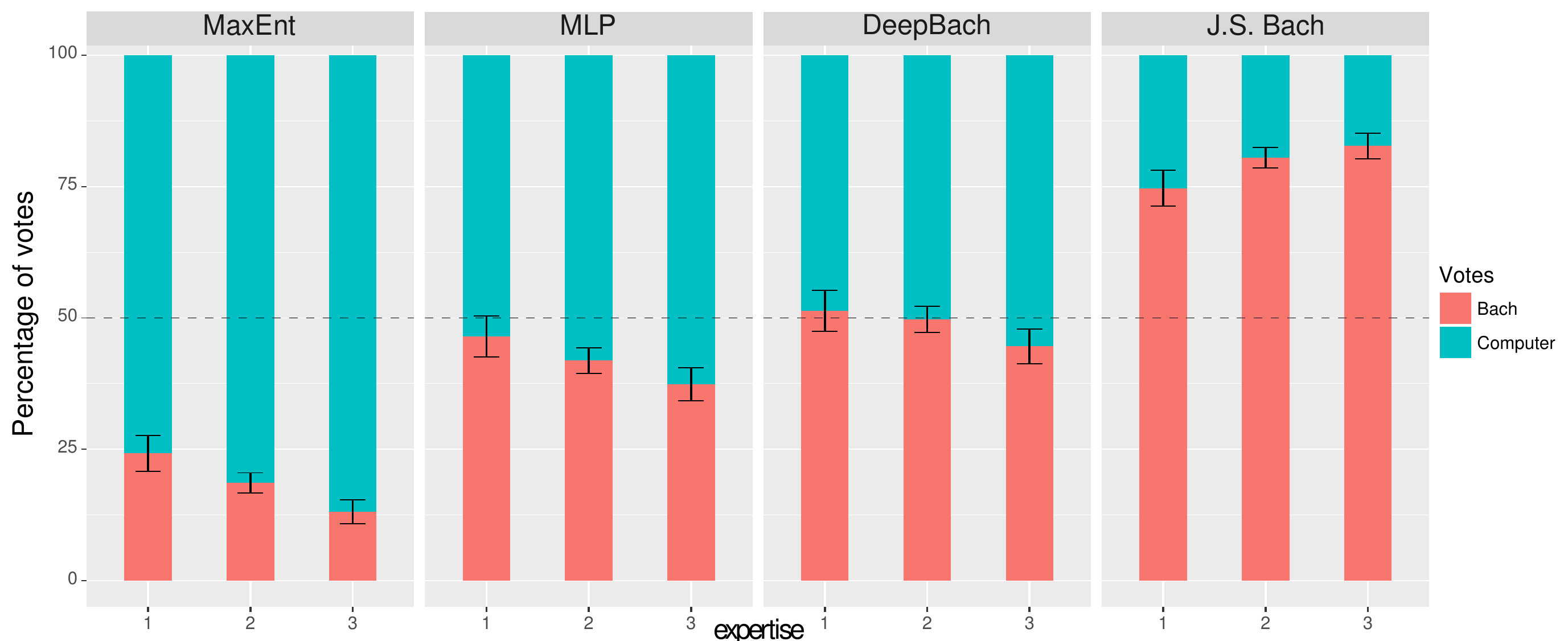}

    \caption{Results of the ``Bach or Computer'' experiment. The figure shows the distribution of the votes between
      ``Computer'' (blue bars) and ``Bach'' (red bars) for each model and each level of expertise of the voters (from 1 to 3), see Sect.~\ref{sec:discrimination-test} for details.}
\label{gameResults}
\end{figure}

\begin{figure}[!h!]

    \centering
    \includegraphics[angle=0, width=0.8 \linewidth]{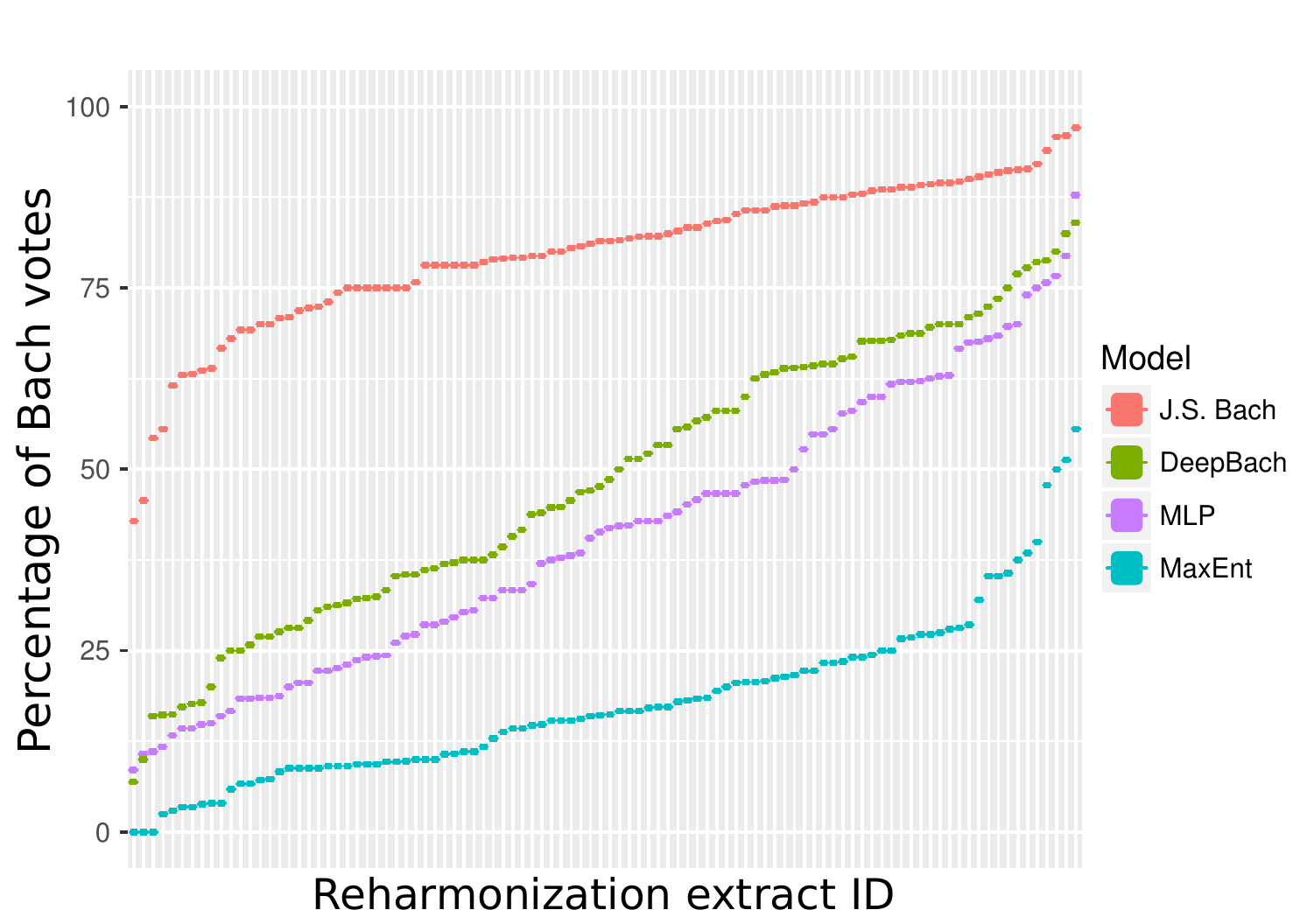}

    \caption{Results of the ``Bach or Computer'' experiment. The figure shows the percentage of votes for Bach for each
      of the 100 extracts for each model. For each model, a specific order for the x-axis is chosen so that the
      percentage of Bach votes is an increasing function of the x variable, see Sect.~\ref{sec:discrimination-test} for
      details.}
\label{gameResultsSongs}
\end{figure}

\begin{figure*}[h!]
    \centering
    \includegraphics[angle=0, width=0.9\linewidth, trim=0 170 0 170, clip]{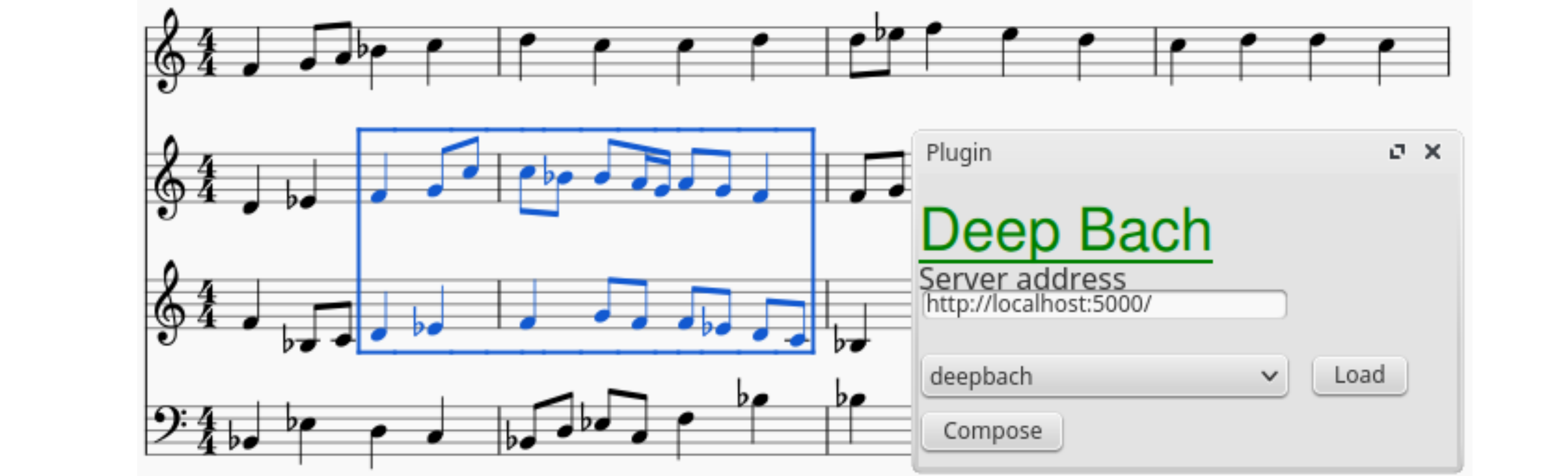}
    \caption{DeepBach's plugin minimal interface for the MuseScore music editor}
\label{fig:editor}
\end{figure*}

\subsection{Discrimination Test: ``Bach or Computer'' experiment}
\label{sec:discrimination-test}
Subjects were presented series of only one musical extract together with the binary choice ``Bach'' or
``Computer''.
% \footnote{This test is available at \url{http://www.flow-machines.com:3010}}
 Fig.~\ref{gameResults} shows how the votes are
distributed depending on the level of musical expertise of the subjects for each model. For this experiment, 1272
people took this test, 261 with musical expertise 1, 646 with musical expertise 2 and 365 with musical expertise 3.

The results are quite clear: the percentage of ``Bach'' votes augment as the model's complexity increase. Furthermore,
the distinction between computer-generated extracts and Bach's extracts is more accurate when the level of musical
expertise is higher.  When presented a DeepBach-generated extract, around 50\% of the voters would judge it as composed
by Bach. We consider this to be a good score knowing the complexity of Bach's compositions and the facility to detect
badly-sounding chords even for non musicians.

We also plotted specific results for each of the 400 extracts. Fig.~\ref{gameResultsSongs} shows for each
reharmonization extract the percentage of Bach votes it collected: more than half of the DeepBach's
automatically-composed extracts has a majority of votes considering them as being composed by J.S. Bach while it is only
a third for the MLP model.

\section{Interactive composition}

\label{sec:interactive}
\begin{figure*}[h!]
  \centering
  \subfloat[]{
    \includegraphics[angle=0, width=0.8\linewidth, trim=0 20 0 20, clip]{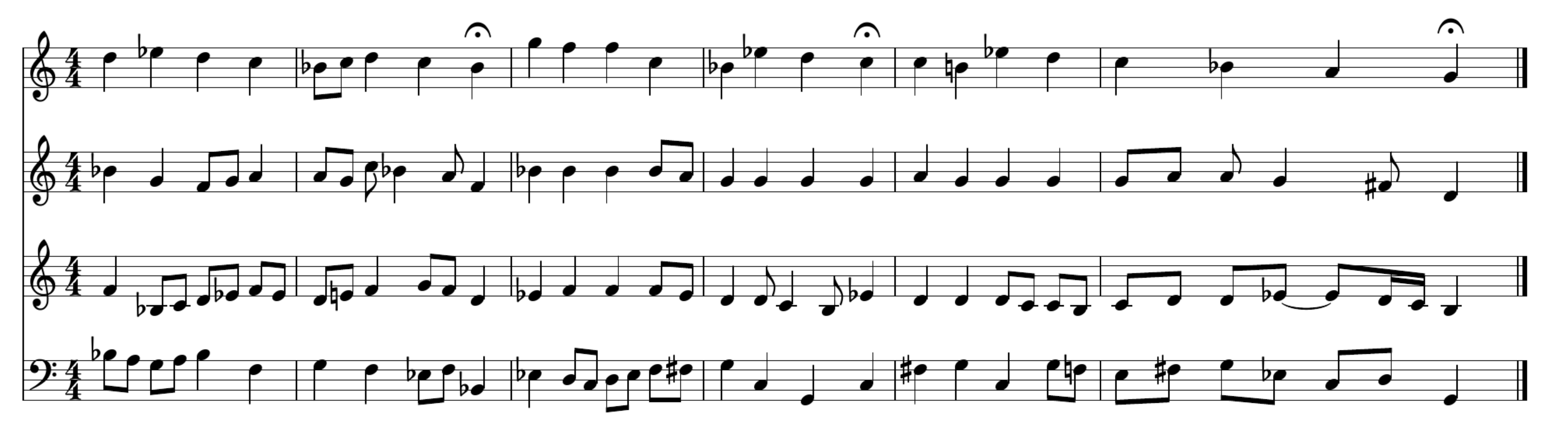}
  } \hfill
    \subfloat[]{
    \includegraphics[angle=0, width=0.98\linewidth]{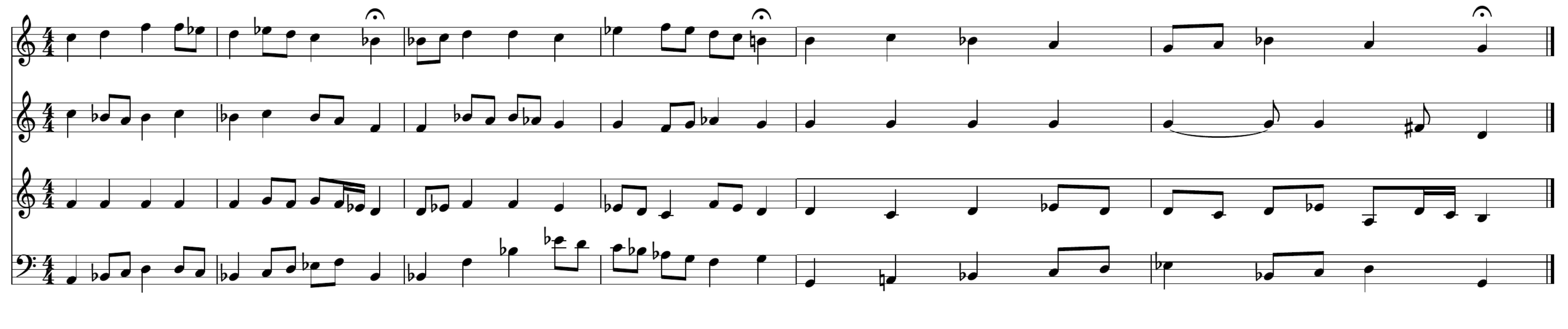}
  }
  \hfill
      \subfloat[]{
        \includegraphics[angle=0, width=0.98\linewidth]{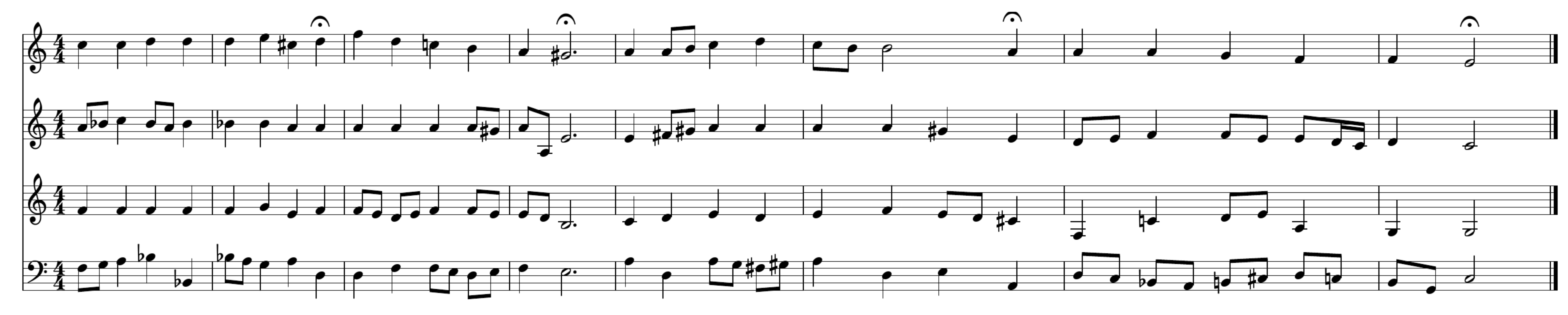}
      }
      \hfill
\caption{Examples produced using DeepBach as an interactive composition tool. Examples (a) and (b) share the same metadata.}
\label{fig:1long}
\end{figure*}

\subsection{Description}

We developed a plugin on top of the MuseScore music editor allowing a user to call DeepBach on any rectangular region. Even if the interface is minimal (see Fig.\ref{fig:editor}), the possibilities are numerous: we can generate a chorale from scratch, reharmonize a melody and regenerate a given chord, bar or part. We believe that this interplay between a user and the system can boost creativity and can interest a wide range of audience. 

\subsection{Adapting the model}
\label{sec:improving}
We made two major changes between the model we described for the online test and the interactive composition tool. 

\subsubsection{Note encoding}
\label{sec:notes}
We changed the MIDI encoding of the notes to a \emph{full name} encoding of the notes. Indeed, some information is lost when reducing a music sheet to its MIDI representation since we cannot differentiate between two \emph{enharmonic notes} (notes that sound the same but that are written differently e.g. F\# and Gb). This difference in Bach chorales is unambiguous and it is thus natural to consider the \emph{full name} of the notes, like C\#3, Db3 or E\#4. 
From a machine learning point of view, these notes would appear in totally different contexts. This improvement enables the model to generate notes with the correct spelling, which is important when we focus on the music sheet rather than on its audio rendering.

\subsubsection{Steering modulations}
We added the \emph{current key signature}  list $\mathcal{K}$ to the metadata $\mathcal{M}$. This allows users to impose modulations and key changes. Each element $\mathcal{K}_t$ of this list
contains the number of sharps of the estimated key for the current bar. It is a integer between -7 and 7. The current key is computed using the key analyzer algorithm from music21.

\subsection{Generation examples}
\label{sec:examples}
We now provide and comment on examples of chorales generated using the DeepBach plugin. Our aim is to show the quality of the solutions produced by DeepBach. For these examples, no note was set by hand and we asked DeepBach to generate regions longer than one bar and covering all four voices.

Despite some compositional errors like parallel octaves, the musical analysis reveals that the DeepBach compositions reproduce typical Bach-like patterns, from characteristic cadences to the expressive use of nonchord tones.
As discussed in Sect.~\ref{sec:improving}, DeepBach also learned the correct spelling of the notes. Among examples in Fig.~\ref{fig:1long}, examples (a) and (b) share the same metadata ($\mathcal{S}, \mathcal{F}$ and $\mathcal{K}$). This demonstrates that even with fixed metadata it is possible to generate contrasting chorales.

 Since we aimed at producing music that could not be distinguished from actual Bach compositions, we had  all provided extracts sung by the Wishful Singing choir. These audio files can be heard on the accompanying website.

\section{Discussion and future work}
\label{sec:disc-future-work}
We described DeepBach, a probabilistic model together with a sampling method which is flexible, efficient and provides
musically convincing results even to the ears of professionals.  The strength of our method is
the possibility to let users impose unary constraints, which is a feature often neglected in probabilistic models of
music. % This relies on Gibbs sampling and a Gibbs sampling-adapted data representation.
Through our graphical interface, the composition of polyphonic music becomes accessible to non-specialists.
The playful interaction between the user and this system can  boost creativity and help explore new ideas quickly.
We believe that this approach could form a starting point for a novel compositional process that could be described as a constructive dialogue between a human operator and the computer.
 This method is general and its implementation simple. It is not only applicable to Bach chorales but embraces a wider range of  polyphonic music. 

 Future work aims at refining our interface, speeding up generation and handling datasets with small corpora.

\bibliographystyle{icml2017}
\bibliography{deepbach}

\end{document}